\documentclass[conference]{IEEEtran}
\usepackage[T1]{fontenc}
\usepackage{times}

\usepackage{tikz}                                                                                                                                                                                    
\usetikzlibrary{shapes,arrows,backgrounds}

\usepackage{cite}
\usepackage[cmex10]{amsmath}
\usepackage{amsfonts}

\usepackage{flushend}

\usepackage{array}

\usepackage{mdwmath}
\usepackage{mdwtab}
\usepackage{fixltx2e}

\usepackage{stfloats}

\usepackage{url}

\begin{document}
%
\title{Multiobjective Tactical Planning under Uncertainty for Air Traffic Flow and Capacity Management}

\author{\IEEEauthorblockN{Ga\'etan Marceau Caron$^{1,3}$}
  \IEEEauthorblockA{$^1$Thales Air Systems\\
    Rungis, France\\
    gaetan.marceaucaron@thalesgroup.com}
  \and
  \IEEEauthorblockN{Pierre Sav{\'e}ant$^2$}
  \IEEEauthorblockA{$^2$Thales Research \&  Technology\\
    Palaiseau, France\\
    pierre.saveant@thalesgroup.com
  }
  \and
  \IEEEauthorblockN{Marc Schoenauer$^3$}
  \IEEEauthorblockA{$^3$TAO project-team, INRIA Saclay Ile-de-France\\
    Universit\'e Paris Sud, Orsay, France\\
    marc.schoenauer@inria.fr}
}

\maketitle

\begin{abstract}
We investigate a method to deal with congestion of sectors and delays in the tactical phase of air traffic flow and capacity management.
It relies on temporal objectives given for every point of the flight plans and shared among the controllers in order to create a collaborative environment.
This would enhance the transition from the network view of the flow management to the local view of air traffic control.
Uncertainty is modeled at the trajectory level with temporal information on the boundary points of the crossed sectors and then, we infer the probabilistic occupancy count.
Therefore, we can model the accuracy of the trajectory prediction in the optimization process in order to fix some safety margins.
On the one hand, more accurate is our prediction; more efficient will be the proposed solutions, because of the tighter safety margins.
On the other hand, when uncertainty is not negligible, the proposed solutions will be more robust to disruptions.
Furthermore, a multiobjective algorithm is used to find the tradeoff between the delays and congestion, which are antagonist in airspace with high traffic density.
The flow management position can choose manually, or automatically with a preference-based algorithm, the adequate solution.
This method is tested against two instances, one with 10 flights and 5 sectors and one with 300 flights and 16 sectors. 



\end{abstract}









\section{Introduction}
Nowadays, delays in air traffic management are a major problem, which is mainly caused by capacity limits, particularly in Europe where the flight density is high.
This relates directly to the regulations taken in Air Traffic Flow and Capacity Management (ATFCM), because their impact on the trajectories are more consequent than the ones taken in Air Traffic Control.
According to \cite{Eurocontrol12}, the Network Manager is responsible for planning the demand issued by the airlines on the air traffic infrastructures, including runways and control sectors. 
This is done in the strategic and the pre-tactical phase, where the different actors are implied in the Collaborative Decision Making process.
Then, the ATFCM daily plan is published, which describes a set of tactical measures, e.g., routeing scenarios or regulations, and the departure slots are assigned to the aircraft operators. 
The latter consist in intervals of 15 minutes that are supposed to encompass the uncertainty at the boarding phase.
Thereafter, the evolution of the flight is assured through the Flow Management Position (FMP), linking every control centers to the Network Manager.
This position is responsible for implementing local procedures, to monitor their effect and to provide the relevant information to the Network Manager.
In return, the Enhanced Tactical Flow Management System is responsible for providing the predicted occupancy count for each sector.
Then, the network manager and the flow management position will agree if the activation of regulations is required or not.

As a matter of fact, the sector occupancy is calculated from a trajectory prediction with potential errors.
Nevertheless, if the predicted sector occupancy is higher than the capacity, regulations can be activated and generate delays.
This will effectively disrupt the initial slot allocation and generate delays.
The major drawback from this procedure is that the effect of regulations becomes effective a few hours later and will drastically impact the workload.
In some cases, when uncertainty about the future is high, ineffective regulations might be issued.
This is the main reason for the introducing of the Short-term ATFCM measures in the process.
These are intended to solve small disruptions locally in time and space, and encompass minor ground delays, flight level capping and minor rerouting.
In the following, we present adaptive target times of arrival as a mean to stabilize the network impacted by small disruptions.

This paper introduces an original methodology to tackle uncertainty regarding aircraft trajectories and airspace sector crossings. 
To propagate the uncertainty, we can infer the probability of sector congestion with a closed-form equation. 
Then, the probabilistic model is used within an optimization algorithm for scheduling all flights at boundaries of the sectors in order to minimize the expected cumulated delays and the expected sector congestion.
To the best of our knowledge, the novelty of this work is to provide the inference mechanism to propagate the uncertainty from the trajectories to the sectors and to use the resulting probabilistic model as a black-box for multiobjective optimization.

The paper is organized as follows: first, we present a literature survey on the formulations and the techniques used to solve the related air traffic flow management problem.
Then, we present the motivation of the paper and the mathematical formulation of the uncertainty model. 
Finally, we give the experimental results that validate the model and the optimisation process, followed by a discussion on the possible extensions of the model and sketches further directions of research.

\section{Related Work}
The Operational Research community has studied many variants of the air traffic flow management problem in the pre-tactical phase since the beginning of the 90s. 
To our knowledge, the most comprehensive formulation is the Air Traffic Flow Management Rerouting Problem \cite{LulliBertsimas2011}, which integrates all phases of a flight, different costs for ground and air delays, rerouting, continued flights and cancellations. 
Instances of the size of the National Airspace of the United States were used to validate the approach. 
Recently, \cite{Clare:2012} recognized the importance of dealing with uncertainty  by minimizing directly the probability of congestion with the concept of chance constraint.
Besides, a multiobjective optimization approach has been used in air traffic control by \cite{Flener2007} to minimize an aggregated complexity metric over sectors, designed and validated by Eurocontrol.
In this study, the dimensions of the multiobjective problem represent the complexity for each sector and thus, they use the weighted sum as a scalarization method to aggregate the complexity over all sectors. 
Consequently, the multiobjective problem is transformed into a mono-objective problem and the weights are used to generate different trade-offs between sectors.
Also, \cite{daniel:2005} use the multiobjective approach to model the trade-off between sector congestion and delays. 
In this case, the objective function space is in two dimensions and the decision space consists of the takeoff time of the flights and the chosen routes.
A multiobjective genetic algorithm is used to generate a pool of solutions with a diversity measure in order to distribute them uniformly on the Pareto front.
From the Optimal Control community, \cite{LeNy2010} proposes a solution to the dynamic problem when weather disruptions occur.
They apply network regulation strategies to a dynamical model representing the problem as a network of queues with load-dependent service rates on sector boundaries.
This macroscopic view, also known as Eulerian Model, is well-adapted for instances of the size of the National Airspace System.
The purpose of the given algorithm is to change the incoming throughput or to reroute flows when a sector capacity degradation occurs.  
Besides, a study of the uncertainty was conducted by \cite{Gilbo11} with an analysis of the prediction error of the time of arrival of the aircraft. 
The main hypothesis of the study is that the random variable of the prediction error follows a Gaussian distribution. 

To the best of our knowledge, this article is the first to tackle the problem of optimization of the air traffic flow and capacity management with a probabilistic model used to monitor the flights in real-time and a multiobjective algorithm to find the adequate actions to respond to disruptions.

\section{Motivation}
The goal of this work is to propose an automated method to enhance the robustness of the network to small disruptions in the tactical phase in ATFCM.
To do so, the temporal uncertainty of flight trajectories are modeled in a dynamical way since the estimated time over (ETOs) can be updated, i.e., via a data-link transmission between ground control and the flight management system of the aircraft or with updates from a ground trajectory prediction.
Through a monitoring process, any change in the sector capacities will result in the evaluation of the congestion from these temporal trajectories. 
If need be, an optimization process is launched to change the trajectories of the aircraft according to the magnitude of the disruption.
For this part, a multiobjective algorithm is used to find multiple plans, which give the tradeoff between generated delays and congestion reduction.
This is a way for the FMP to be able to choose manually or automatically the most adapted solution to the disruption.

The main novelty of this paper is to consider the uncertainty at a trajectory level by the intermediate of the target time of arrival on the waypoints, then to propagate the uncertainty into the sectors by using a closed-form equation and finally, to use the probability marginals in order to infer the expected cost of delays and the expected cost of congestion.
Then, the targets are tuned with a multiobjective algorithm to reduce the expected cost of congestion, if it is higher than a threshold.
Also, we include ground delays since the actions on airborne flights are clearly restricted by justified economic reasons from the airlines. 
 

A dynamic and stochastic approach can generate a plan that is robust to changes as long as the uncertainty is well-estimated.
An optimal schedule of a deterministic approach is characterized by the tightness between the target times of arrival.
To minimize the delays, it should imply that an aircraft enters in the sector as soon as another aircraft leaves it.
The difference in time between these targets will be of the same order than the time discretization, e.g., one minute. 
With a probabilistic approach, we use safety margins that are function of the presence of uncertainty in the system.
The main difference with robust approaches, which considers the worst-case scenario, is that the plan is not too conservative. 
In effect, the probability that the worst case arises might be so low that it will lead to a suboptimal behavior. 
Instead, it is more interesting to consider the scenarios proportionally to their probability of occurrence.

Applying multiobjective optimization in this context is a way to gather multiple schedules, each corresponding to a trade-off between minimizing the use of regulations and reducing the complexity. 
Here, this latter refers to the probabilistic occupancy count of a sector, which is a refinement of the deterministic one used today.
A monitoring process raises an alarm when the occupancy count reaches 90\% of the capacity at any time. 
Since it is a scalar value for the entire sector, it does not account for the geometries of the trajectories. 
As a matter of fact, increasing the number of flights by one in the sector might increase drastically the workload of the controllers depending on the current airspace. 
If the traffic is organized, the increase will be small, but if the flights have many crossing trajectories or have any flight level changes, the disruption might be significant.
Consequently, the occupancy count is not sufficient alone to evaluate the impact of the number of flights on the workload of the controllers and so, additional air traffic control tools are necessary to manage locally and spatially the trajectories.

\section{Mathematical Formulation}

\subsection{Notations}
Essentially, the mathematical formulation will rely on the probability theory and stochastic optimization.
Here, the events are target times of arrival at georeferenced points.
Let us consider flight plan $f \in \mathcal{F}$ with $n$ waypoints denoted by $X_1^f, \dots, X_n^f$  and associated to $n$ random variables $T_1^f, \dots, T_n^f$, where $T_i^f$ represents the time of overfly of flight $f$ over waypoint $X_i^f$, and let us call $p_i^f$ the probability density of $T_i^f$, dropping the superscript $f$ when there is no ambiguity.
According to standard definition, the marginal probability is $\Pr \left[ T_i \in \Delta t \right] = \int_{\Delta t} p_i(\tau) \, d\tau$.
This simply refers to the probability for a flight to be over $X_i$ during the time interval $\Delta t$.
Also, because we consider the flight plan as an ordered sequence, the joint probability density function from point $X_i$ to $X_j$ is expressed by $p_{i:j}(t_{i:j})$, where $t_{i:j} \in \mathbb{R}^{j-i+1}$ is a vector with time values for every points of the sub-sequence.
Finally, the conditional probability is: 
\begin{equation}
  \Pr \left[T_j \in \Delta t| T_i = t_i \right] = \int_{\Delta t} p_{j|i}(\tau|t_i) d\tau
\end{equation}
where $p_{i|j}(\tau|t_j)$ is the conditional probability that the flight is over $X_j$ during the time interval $\Delta t$ given that the flight is over the point $X_i$ at time $t_i$.

\begin{figure}
  \centerline{
    \begin{tikzpicture}
      [scale=0.6,every node/.style={circle,fill=green!20,scale=0.7}]
      \node (n11) at (0,2) {$T_1^1$};
      \node (n12) at (2,2)  {$T_2^1$};
      \node (n13) at (4,2) {$T_3^1$};
      \node (n14) at (6,2) {$T_4^1$};
      \node (n21) at (0,-2) {$T_1^2$};
      \node (n22) at (2,-2)  {$T_2^2$};
      \node (n23) at (4,-2) {$T_3^2$};
      \node (n24) at (6,-2) {$T_4^2$};
      \node (t11)[rectangle,fill=red!20] at (0,4) {$\gamma_1^1$};
      \node (t12)[rectangle,fill=red!20] at (2,4) {$\gamma_2^1$};
      \node (t13)[rectangle,fill=red!20] at (4,4) {$\gamma_3^1$};
      \node (t14)[rectangle,fill=red!20] at (6,4) {$\gamma_4^1$};
      \node (t21)[rectangle,fill=red!20] at (0,-4) {$\gamma_1^2$};
      \node (t22)[rectangle,fill=red!20] at (2,-4) {$\gamma_2^2$};
      \node (t23)[rectangle,fill=red!20] at (4,-4) {$\gamma_3^2$};
      \node (t24)[rectangle,fill=red!20] at (6,-4) {$\gamma_4^2$};
      \node (ns1)[diamond,fill=blue!20] at (1,0) {$S_1$};
      \node (ns2)[diamond,fill=blue!20] at (3,0) {$S_2$};
      \node (ns3)[diamond,fill=blue!20] at (5,0) {$S_3$};
      \draw[->] (t11) edge (n11);
      \draw[->] (t12) edge (n12);
      \draw[->] (t13) edge (n13);
      \draw[->] (t14) edge (n14);
      \draw[->] (t21) edge (n21);
      \draw[->] (t22) edge (n22);
      \draw[->] (t23) edge (n23);
      \draw[->] (t24) edge (n24);
      \draw[->] (n11) edge (n12);
      \draw[->] (n12) edge (n13);
      \draw[->] (n13) edge (n14);
      \draw[->] (n21) edge (n22);
      \draw[->] (n22) edge (n23);
      \draw[->] (n23) edge (n24);
      \draw[->] (n11) edge (ns1);
      \draw[->] (n21) edge (ns1);
      \draw[->] (n12) edge (ns1);
      \draw[->] (n22) edge (ns1);
      \draw[->] (n12) edge (ns2);
      \draw[->] (n13) edge (ns2);
      \draw[->] (n22) edge (ns2);
      \draw[->] (n23) edge (ns2);
      \draw[->] (n13) edge (ns3);
      \draw[->] (n23) edge (ns3);
      \draw[->] (n14) edge (ns3);
      \draw[->] (n24) edge (ns3);
    \end{tikzpicture}
  }
  \caption{\label{fig:bn}Bayesian Network for a flight plan}
\end{figure}
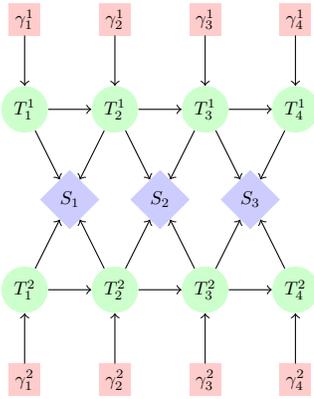

\subsection{Trajectory Model}
Let's define an uncertainty model for any trajectory.
Figure \ref{fig:bn} gives a simple Bayesian Network where an arrow between $T_i$ and $T_{i+1}$  shows that the former influences the latter, or more precisely, that the two random variables are not independent.
The joint density function of $T_i$ and $T_{i+1}$ is:
\begin{eqnarray}
\label{eq:joint}
p_{i,i+1}(t_i,t_{i+1}) &=& p_{i+1|i}(t_{i+1}|t_i) \cdot p_i(t_i)
\end{eqnarray}
This equality represents the propagation of the information in the same direction than the sequence of waypoints.
As a first physical constraint, in order to respect the arrow of time along the sequence, we impose:
$$p_{i,j}(t_i,t_j) = 0, \; \mbox{ if } t_i \ge t_j, \; \forall j > i$$

Now, let's generalize the joint distribution for an arbitrary number of waypoints:
\begin{eqnarray}
\label{eq:fpModel}
p_{1:N}(t_{1:N}) &=& \prod_{i=2}^N p_{i|i-1}(t_i|t_{i-1}) p_1(t_1)
\end{eqnarray}
This is the markovian uncertainty model for the flight plan.
On the one hand, $p_1(t_1)$ is the density function associated to the arrival of the flight in the airspace.
On the other hand, $p_{i|i-1}(t_i,t_{i-1})$ is the density function, which contains information on the intents of the pilot to arrive at a point given the time of arrival on the previous points. 

\subsection{Sector Model}
Here, we give the closed-form equation for computing the exact probability that a sector is congested, which requires $S_{s,f}^t$, the Bernoulli random variable that the flight $f$ is in the sector $s$ at time $t$, and $\overline{S_{s,f}^t}$ its complementary.
Notice that $\{ S_{s,f}^t : t \in \Omega \}$ is a stochastic process where $\Omega$ is the time horizon.

Then, the probability to {\bf not} be in the sector during the time interval $\Delta t=[t_{min},t_{max}]$ is the probability to enter after $t_{max}$ or the probability to exit before $t_{min}$.
Because of the arrow of time constraint, these two events are mutually exclusive and one obtain:
\begin{eqnarray}
\label{eq:probSector}
  \Pr(\overline{S_{s,f}^{\Delta t}}) &=& \Pr(T_i^f > t_{max}) + \Pr(T_j^f \le t_{min}) \nonumber \\
  &=& \left[ 1 - \Pr(T_i^f \le t_{max}) \right] + \Pr(T_j^f \le t_{min}) \nonumber \\
  &=& 1 - F_i^f(t_{max}) + F_j^f(t_{min}) \nonumber \\
  \implies \Pr(S_{s,f}^{\Delta t}) &=& F_i^f(t_{max}) - F_j^f(t_{min})
\end{eqnarray}
When $(t_{max} - t_{min}) \rightarrow 0$, we obtain the values for $S_{s,f}^t$.

Now, inference on the presence of many flights in a given sector during an interval can be undertaken.
To do so, let $K_s^t$ be the random variable of the number of flights in the sector $s$ at time $t$.
Then, by using a multi-index notation, we have:
\begin{equation}
\label{eq:cong}
\Pr(K_s^t = n) = \sum_{|a|=n} \prod_{f \in \mathcal{F}} \Pr(S_{s,f}^t)^{a_f} \cdot \Pr(\overline{S_{s,f}^t})^{1-a_f}
\end{equation}
where $a = \left( a_1, a_2, \dots, a_{N_s^t} \right) \in \{0,1\}^{N_s^t}$, $|a| := a_1 + \dots + a_{N_s^t}$ and $N_s^t=| \left\{ i|\Pr(S_{s,i}^t) \ne 0 \right\}|$.
Again, $\{ K_s^t : t \in \Omega\}$ corresponds to a stochastic process and these are depicted with diamonds on the graphical model.

\subsection{Poisson Binomial Distribution}
Since \cite{Wang93}, we know that the congestion model (eq. \ref{eq:cong}) is a Poisson Binomial distribution.
This model is close to the one proposed by \cite{Gilbo11}, except that they are interested by the parameters of a Gaussian distribution around the error of prediction to correct the occupancy count.
Here, we are interested in the expected number of aircraft in the sector, and so, we need to compute the probability mass function.
As an example, if we consider three flights, the equation becomes:
\begin{eqnarray}
\Pr(K_s^t = 1) &=& \Pr(S_{s,1}^t)\Pr(\overline{S_{s,2}^t})\Pr(\overline{S_{s,3}^t}) \nonumber \\ 
&+& \Pr(\overline{S_{s,1}^t})\Pr(S_{s,2}^t)\Pr(\overline{S_{s,3}^t}) \nonumber \\ 
&+& \Pr(\overline{S_{s,1}^t})\Pr(\overline{S_{s,2}^t})\Pr(S_{s,3}^t) \nonumber
\end{eqnarray}
As a first remark, the number of conjunctions (products) is determined by the number of combinations $\binom{N}{n}$ where $N$ is the total number of flights crossing the sector and $n$ is an arbitrary number of flights. 
Consequently, the associated computational burden attains its maximum value at $n = N/2$ and decreases when $n$ goes to $0$ or $N$.
Consequently, we need to rely on the characteristic function, as demonstrated by \cite{Hong11oncomputing}.
From this work, we know that the probability can be expressed by:
\begin{eqnarray}
\label{eq:cong2}
\Pr(K_s^t = n) &=& \frac{1}{N_s^t+1} \sum_{l=0}^{N_s^t} \exp(-iwln) \cdot \nonumber \\
& & \prod_{f=1}^{N_s^t} \left[ 1-\Pr(S_{s,f}^t)+\Pr(S_{s,f}^t) \exp(iwl) \right] \nonumber
\end{eqnarray}
where $i=\sqrt{-1}$ and $w=\frac{2 \pi}{N_s^t+1}$. 
Therefore, this equation can be computed efficiently with the use of a Fast Fourier Transform.

\subsection{Flight Intents}
At this point, one would like to manipulate the flight intents more directly, i.e., for the generation of the conditional probabilities and during the optimization process. 
To do so, let $\gamma_{i+1}$ be the target time of arrival on the waypoint $X_{i+1}$ of an arbitrary flight.
Now, we make the assumption that the flights have a unique target time of arrival on each waypoint.
Then, the conditional probability can be parameterized as $p_{i+1|i}(t_{i+1}|t_i;\gamma_{i+1})$.
An acceptable constraint on the space of possible conditional probabilities is to restrain it to unimodal functions where their mode is centered at the target value. 
Furthermore, we require that its support is bounded to denote the physical constraints of the aircraft, i.e. its flight envelope and its finite amount of fuel.
Good candidates for such properties are triangular and gamma probability density functions, already used in project management tools, like PERT, for characterizing the length of a task in a scheduling problem.
Then, as depicted on the graphical model, the rectangular nodes act as interfaces between the optimization algorithm and the model for computing the expected cost functions.
This corresponds to a black-box optimization approach.

\subsection{Optimization Formulation}
Now, we have all the elements to define our multiobjective optimization problem. 
Because of the stochastic context, one way to define the cost functions is to use their expected values. 
For the expected cost of delays, let $\phi_f:\Omega \rightarrow \mathbb{R}_+$ be the cost of delays function for the flight $f$. 
Then, the expected cost of delays for this flight is:
$$\mathbb{E}_{\phi_f}(T_{n_f}^f;\gamma_{|f}) = \int_{\Omega} \phi_f(\tau) \cdot p_{n^f}^f(\tau;\gamma_{|f}) d\tau$$
where $n_f$ is the number of waypoints in the flight plan $f$ and so, $p_{n^f}^f$ refers to the marginal density function associated to the arrival point $X_{n^f}^f$.
The inequality ensures that the cost function is bounded for the values in the support of the probability density function.
When aggregating these individual functions in order to obtain the associated objective function, one question that immediately arises is equity.
In this work, we define the same cost function for every flight and use the super-linear trick from \cite{LulliBertsimas2011}, in order to penalize exponentially any delays. 
As a consequence, we avoid the case where a flight will be constantly penalized for the benefit of the others.
So, we use $\phi(\tau) = (\tau - A_f)_+^\beta$ where $A_f$ is the scheduled time of arrival of flight $f$, $\beta > 1$ is the super-linear coefficient and the plus index refers to the positive part.
One can also find other relevant cost functions without changing the optimization formulation.
In our work, we use: 
\begin{equation} 
\label{eq:cost1}
\mathcal{C}_1(\gamma) = \sum_{f \in \mathcal{F}} \left[ \int_{\Omega} (\tau - A_f)_+^2 \cdot p_{n^f}^f(\tau;\gamma_{|f}) d\tau \right]
\end{equation}
as the first objective. 
Here, $\gamma_{|f}$ denotes the vector of intents restricted to the ones concerning the flight $f$.
Notice that $p_{n^f}^f(\tau;\gamma_{|f})$ is the resulting marginal density function obtained from marginalizing the joint probability distribution obtained with eq. \ref{eq:joint} where all the components of $\gamma_{|f}$ are implied in the propagation of the uncertainty.
Because $\mathcal{F}$ is finite and the support of $p_{n^f}^f$ is bounded, we know that $\mathcal{C}_1$ is finite for any targets.

In the same manner, we define the cost of congestion of sector $s$ by $\psi_s: \mathbb{N} \times \Omega \rightarrow \mathbb{R}_+$ with the number of flights and the time as arguments.
The expected cost of congestion is: 
$$\mathbb{E}_{\psi_s(\cdot,t)}(K_s^t) = \sum_{n=C_s^t+1}^{N_s^t} \psi_s(n,t) \cdot \Pr(K_s^t=n)$$
where $C_s^t$ is the capacity of the sector $s$ at time $t$.
Let $\bar{\Omega}$ be the temporal interval from now to the upper bound of the support.
\begin{equation}
\label{eq:cost2}
\mathcal{C}_2(\gamma) = \int_{\bar{\Omega}} \sum_{s \in \mathcal{S}} \sum_{n=C_s^\tau+1}^{N_s^\tau} (n - C_s)^{\lambda} \cdot \Pr(K_s^\tau=n;\gamma) d\tau \nonumber
\end{equation}
Again, $\Pr(K_s^t=n;\gamma)$ is the resulting probability distribution of the inference done with equation \ref{eq:cong}, which depends on the intents $\gamma$.
Here, the parameter $\lambda$ denotes the risk aversion of the controllers when exceeding the capacity.
Because $\mathcal{S}$ is finite and the stochastic process will eventually converge toward zero, $\mathcal{C}_2$ is finite for any targets.

$\mathcal{C}_1$ and $\mathcal{C}_2$ are the two criteria of our bi-objective optimization problem.
Let $\mathcal{D} \subseteq \mathbb{R}^n$ be the decision space and $f:\mathcal{D} \rightarrow \mathbb{R}^2$ be the vector-valued cost function.
Let $x \in \mathcal{D}$ be a point in our decision space, each dimension of $f(x) = (\mathcal{C}_1(x),\mathcal{C}_2(x))$ denotes a cost associated to the decision.
When there is a heavy demand on the airspace, the two costs are antagonist, i.e., reducing the delays will induces more flights in the airspace and, as a consequence, will increase the congestion probability.
This idea is captured by the relation of Pareto dominance.
Let $x,y \in \mathcal{D}$ be two decisions, then $x$ dominates $y$, denoted by $x \prec y$, iff $\mathcal{C}_i(x) \le \mathcal{C}_i(y), \; \forall i \in \left\{ 1,2 \right\}$ and $\exists j \in \left\{ 1,2 \right\} \;|\;  \mathcal{C}_j(x) < \mathcal{C}_j(y)$.

\subsection{Constraints}
From the optimization algorithm point of view, the intents shall be bounded with the flight envelope.
However, assigning a value to a target will impact the bounds of the subsequent targets. 
To obtain independent decision variables, the optimization algorithm works with flight durations in sectors.
These bounds are hard constraints, which cannot be violated in order to find better solutions and define feasible intervals. 
There is a distinction between feasible intervals and probable intervals defined by the supports of the marginal distributions. 
So, for a given point, a probable interval must be a subset of the feasible interval.
Therefore, we only consider box constraints $\gamma_i \in \left[\underline{\gamma_i} , \overline{\gamma_i} \right]$, which are easily taken into account in evolutionary algorithms in general.
As in \cite{Flener2007}, an en-route flight can have a maximum of speed up rate of 1 minute per 20 minutes and a maximum slow down rate of 2 minutes per 20 minutes.
During experimentations, we notice that en-route speed changes are insufficient to reduce the cost of congestion, considering that the mean duration in sectors is around 20 minutes. 
Also, we add the possibility to increase the distance in sectors by a factor $\delta$ and we introduce the possibility for ground delays.
So, we obtain the intervals:
\begin{equation}
\label{eq:recursive}
 \gamma_i \in \left[ \frac{\beta \cdot V_{i,i+1}}{d_{i,i+1}}, \frac{\alpha \cdot V_{i,i+1}}{\delta \cdot d_{i,i+1}} \right]
 \end{equation}
 where $V_{i,i+1}$ is the mean speed and $d_{i,j}$ is the orthodromic distance between $i$ and $i+1$.
In this study, we choose $\alpha = 0.9$ and $\beta=1.05$ and $\delta=1.05$.

\subsection{Monte-Carlo Simulation}
In order to verify experimentally the inference equations, we rely on Monte-Carlo approximation.
First, we need to be able to sample from the flight model of each flight (see eq. \ref{eq:fpModel}).
Because of the structure of our graphical model, a simple forward sampling technique can be used with the conditional probabilities of the model.
A sample is expressed as a vector $x_f^{(k)} \sim p_{1:N}^f(t_{1:N})$ with a time of overflight for each point of the flight $f$. 
Then, we can construct an indicator function per flight to represent the fact that the flight is in the sector at time $t$.
We will denote these by $s_f^{(k)}(t) = \chi_{[x_{f,in}^{(k)},x_{f,out}^{(k)}]}(t)$, where $\chi$ is the indicator function.
Then, we can count the number of flight in the sector for any time: $s^{(k)}(t) = \sum_{f \in \mathcal{F}} s_f^{(k)}(t)$.
To determine if the sector is congested, we use again the indicator function with the set $C$ of values higher than the capacity: $c^{(k)}(t) = \chi_C(s^{(k)}(t)$.
Finally, the Monte-Carlo approximation is given by: $c_{\mathcal{K}}(t) = \frac{1}{|\mathcal{K}|} \sum_{k \in \mathcal{K}} c^{(k)}(t)$ where $\mathcal{K}$ is the set of Monte-Carlo simulations.
At this point of the research, the proof of convergence of the Monte-Carlo Routine toward the inference equation for the probability of congestion is unknown. 
The main difficulty comes from the fact that the closed-form equation concerns one timestamp at a time instead of the Monte-Carlo routines, which generete time intervals.
Therefore, in this study, we solely verify experimentally that the two methods return the same result.

\section{Experiments}
In this section, we describe the experiment on reducing the congestion with a multiobjective optimization algorithm and a probabilistic model.
The first goal is to numerically validate the theoretical model defined above, and to assess the propagation of the uncertainty from the trajectories to the sector. 
The second goal is to assess whether NSGA-II can actually solve the multiobjective optimization problem.
According to \cite{Deb00afast}, the chosen variation operators, simulated binary crossover operator (SBX) and the polynomial mutation, are well suited for bounded continuous design variables.
Thereafter, we choose a population size of 100 individuals, a crossover probability of 0.8 with an expanding coefficient of 20 for the SBX operator, a mutation probability of 0.2 and an expanding coefficient of 20 for the polynomial mutator.
Also, the maximum number of generations allowed is 400.



\begin{figure}
  \centering
  \includegraphics[scale=0.28]{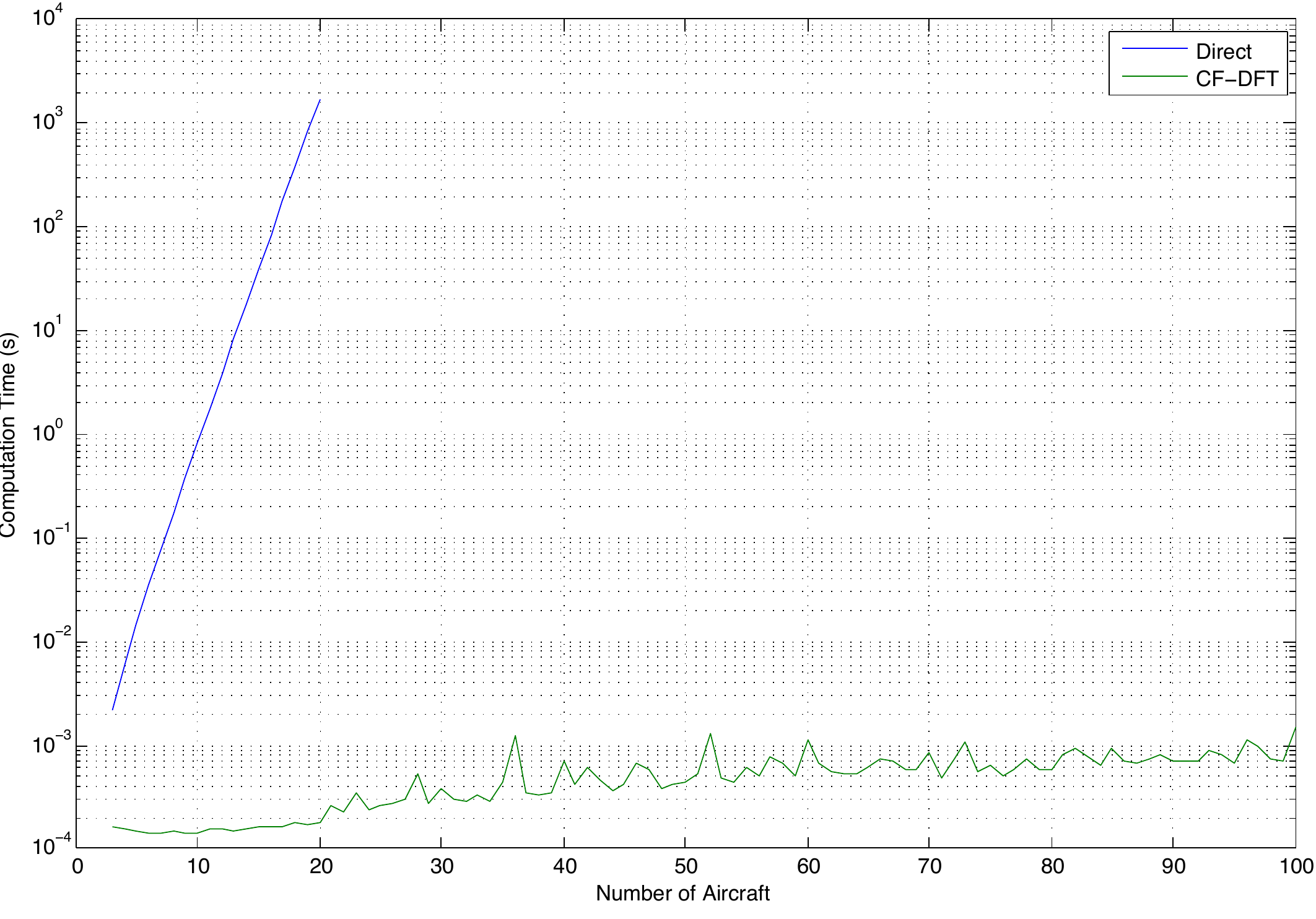}
  \caption{\label{fig:computation_time} Comparative of computation time for Poisson Binomial}
\end{figure}

\subsection{Assumptions}
First of all, we discretize the temporal horizon of the probabilistic model in order to compute numerically the integrals.
We choose a time step of one minute because we believe that it is under the order of magnitude of the precision in real world.
This choice affects the accuracy of the evaluation of the uncertainty and therefore and can be used to control the trade-off between accuracy and computational burden.
Thereafter, we assume that the feasible interval length for the first point is from 5 minutes to 60 minutes and the support length of the probability distribution cannot be less than 15 minutes.
For the next points, we consider solely flights with constant level flight above FL300 with a true airspeed of 460 knots (MACH 0.78).
With $\alpha=0.9$ and $\beta=1.05$, the flight can lower its speed to 414 knots (MACH 0.72) and increase it to 483 knots (MACH 0.82).
For the conditional probability, we assume that the flight management system of the aircraft tries to maximize the probability to arrive at the next point at the given target.
A way to encode it is to take a probable interval and to map a triangular distribution over it where the intent is the mode of the distribution.
In our case, we choose a probable interval length equals to a confidence interval of 95\% on the distribution obtained in \cite{Gilbo11}, that is a length of 24 minutes.

\subsection{Experimental Setting}
The chosen instance for this study, implies 10 flights and 5 sectors disposed as a X, with a central sector that every flight must cross. 
The capacities are three flights for the central one, two for the northwest and southeast ones and one for the southwest and northeast ones.
The underlying decision space consists in a 60 dimensional space with the departure nodes.
A second instance was used to assess the performance of the inference algorithm in the probabilistic algorithm.
The instance comprises 300 flights, regrouped in 10 flows, each crossing 4 sectors. These are defined on a 4x4 grid and the flows arrives from the north and the east, including the diagonals from northwest and southeast.
The model was coded in C++ and simulated on a 2.2GHz processor.
For the algorithm using the characteristic function, we used the FFTW library \cite{Frigo05thedesign}.
The time required to simulate the model, i.e., for a function evaluation, is around 113ms with the direct method for the first instance and 3 seconds for the second with the FFT algorithm.
As a matter of fact, we can see from on the log-scale of figure \ref{fig:computation_time} that the FFT algorithm is clearly faster than the direct one.
For NSGA-II, we used the Paradiseo Library also in C++.
This solver can be freely downloaded on the INRIA's forge at : \url{http://paradiseo.gforge.inria.fr/}.
In order to assess the performance of the algorithm, we used the hypervolume indicator implemented in PISA \cite{KTZ2005a}.
To verify the model, we used the Monte-Carlo routine defined previously. 
The associated results are depicted on the different figures with crosses. 

\begin{figure}
  \centering
  \includegraphics[scale=0.35]{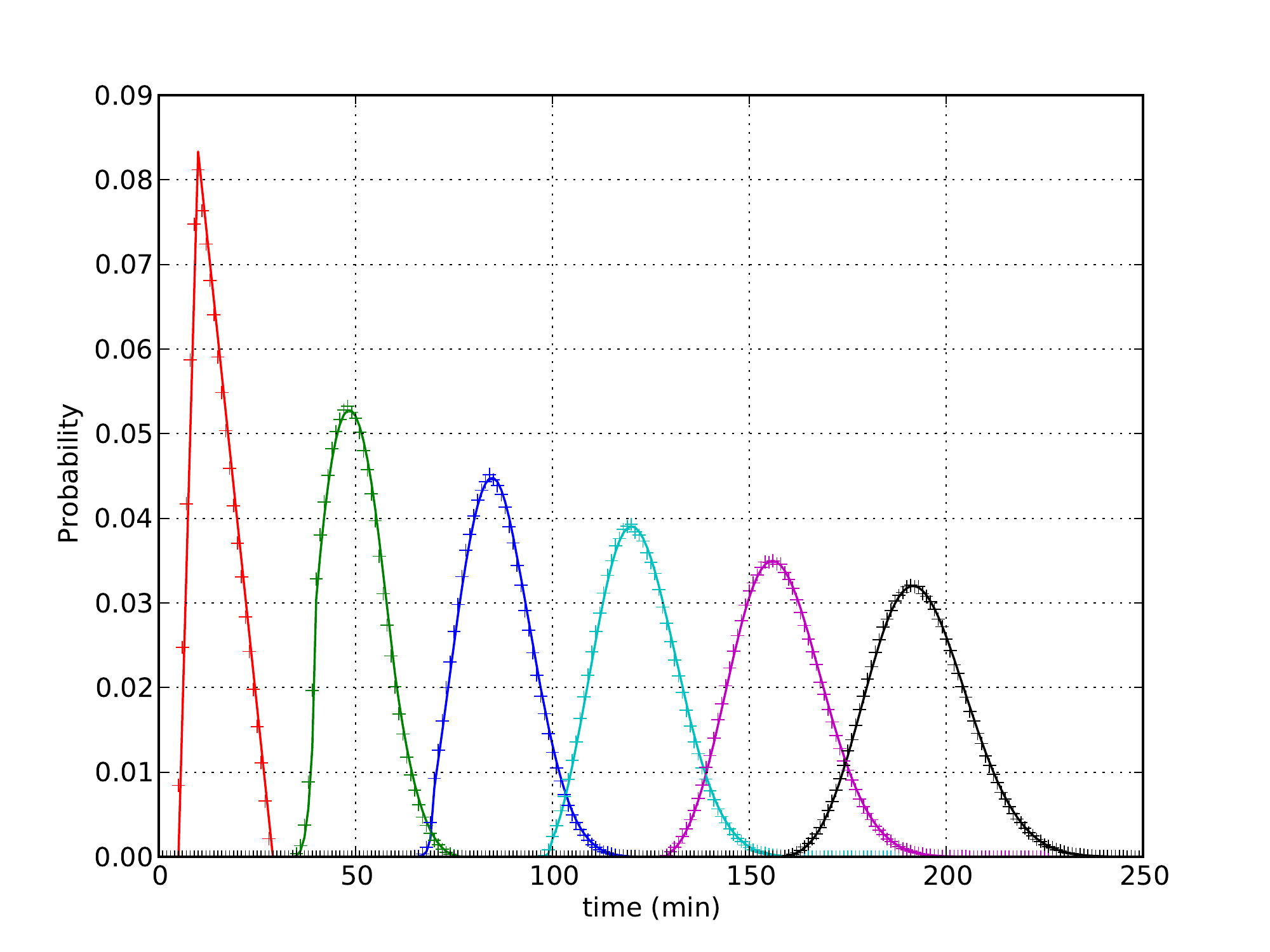}
  \caption{\label{fig:marginals} Marginals probability over five waypoints}
\end{figure}

\begin{figure}
  \centering
  \includegraphics[scale=0.35]{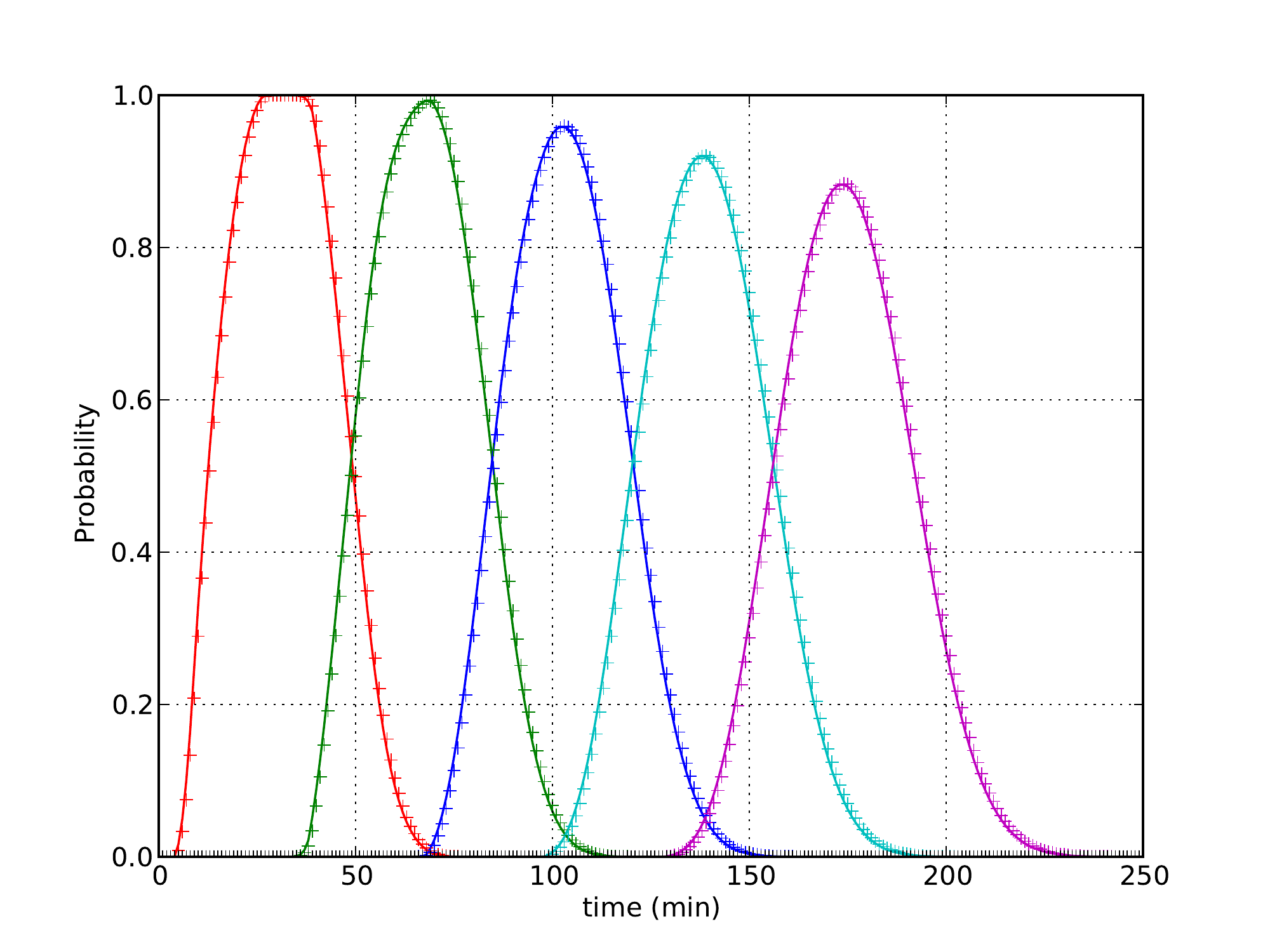}
  \caption{\label{fig:probSector}Probability to be in different sectors}
\end{figure}

\subsection{Analysis}
Now, we will analyze each step of the methodology.
First, we need to compute the marginal probabilities over the waypoints, as depicted on Figure \ref{fig:marginals}, with the trajectory model defined by equation \ref{eq:fpModel}.
We can see that the value of the modes decreases and the supports of the distributions increase with time. 
This simply translates the fact that uncertainty on the target time of arrival increases with time. 
Then, we notice that the mean difference between the modes of the marginals is around 7 minutes higher than the expected 30 minutes required for crossing a sector. 
As a matter of fact, when the lower bound of the probable interval is lower than the feasible interval, the probability is skewed with a heavy tail toward the future, i.e., there is more room for losing time with the coefficient on speed and on distance. 


\begin{figure}
  \centering
  \includegraphics[scale=0.35]{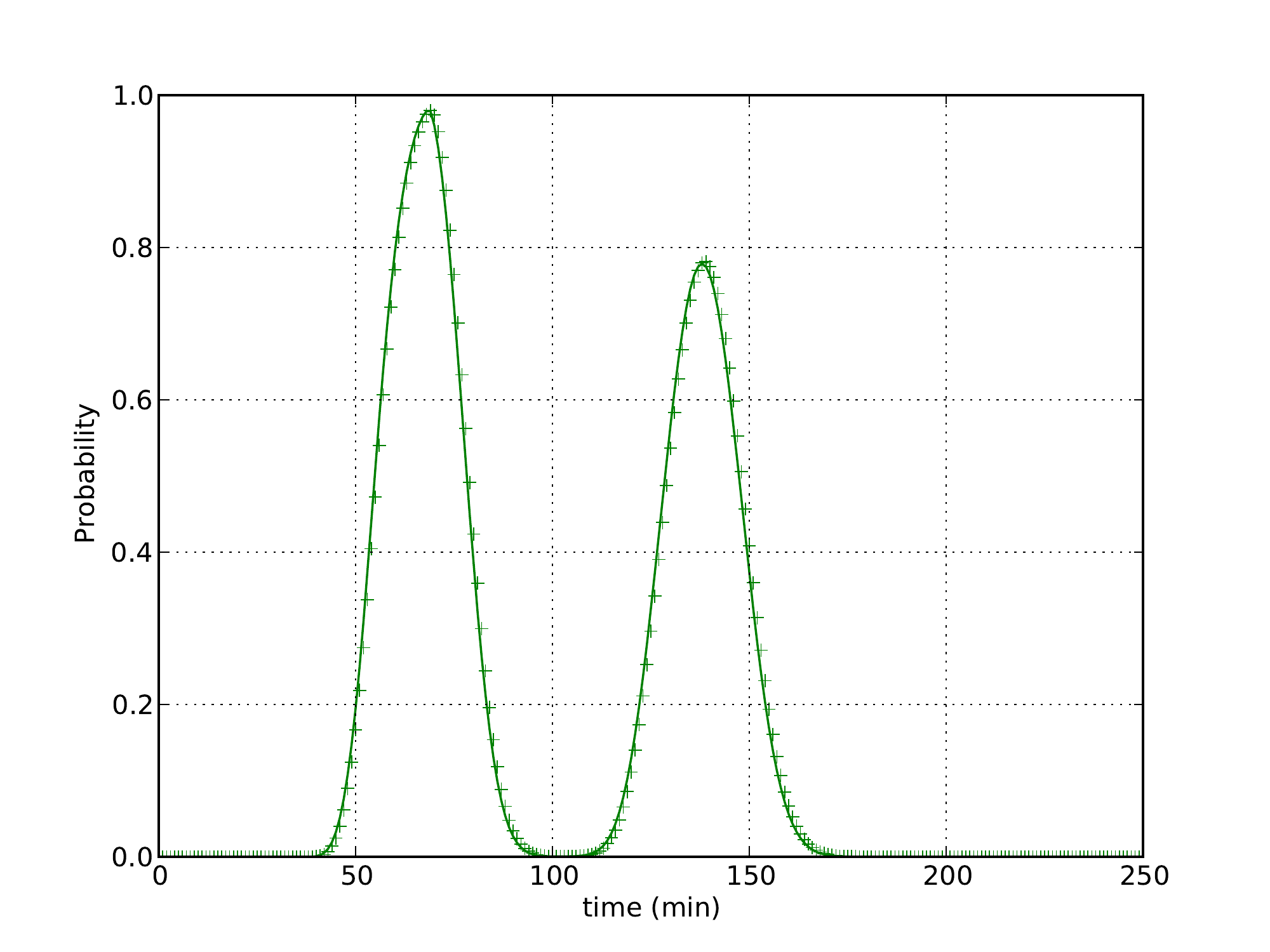}
  \caption{\label{fig:congestion}Probability of congestion}
\end{figure}

Then, with equation \ref{eq:probSector}, we compute the stochastic process for a flight to be in a sector over time, as depicted on Figure \ref{fig:probSector}.
We can see that the values of modes decrease with time.
Also, there is more uncertainty when the coordination between sectors occurs, that is to say the overlapping regions of the different curves grow in time.
We can also visualize the fact that the probability for an aircraft to be in one of the sectors is equal to 1 by summing the probability at every timestamp.
Besides, we can expect that the function defining the stochastic process will be unimodal since there is no reason for the probability to vary while the aircraft is inside the sector.

Thereafter, we need to compute the probability that the sector is congested at a given time, using equation \ref{eq:cong}.
Figure \ref{fig:congestion} shows the stochastic process associated to the northeast sector.
Again, even if the instance is symmetric in distance, the probability for this sector to be congested decreases with time since the variance of the marginals increases with time.
It means that some aircraft will cross the sector before the others without any regularization.
As a matter of fact, this model is a way to quantify the effectiveness of regularizations. 
Combined with an optimization algorithm, we are able to put priority on actions affecting events that will occur before others, or at least, on events with more confidence of occurrence without the use of a discount factor.

When all these distributions are known, the probabilistic model can compute the expected cost of delays and congestion using equations \ref{eq:cost1} and \ref{eq:cost2} respectively.
One way to understand the cost functions from a computational point of view is to add, for all possible timestamps on the temporal horizon, the cost function at a given timestamp multiplied by the probability at this timestamp. 
Consequently, minimizing the probability of congestion for every timestamp will effectively minimize the expected cost of congestion.

\begin{figure}
  \centering
  \includegraphics[scale=0.35]{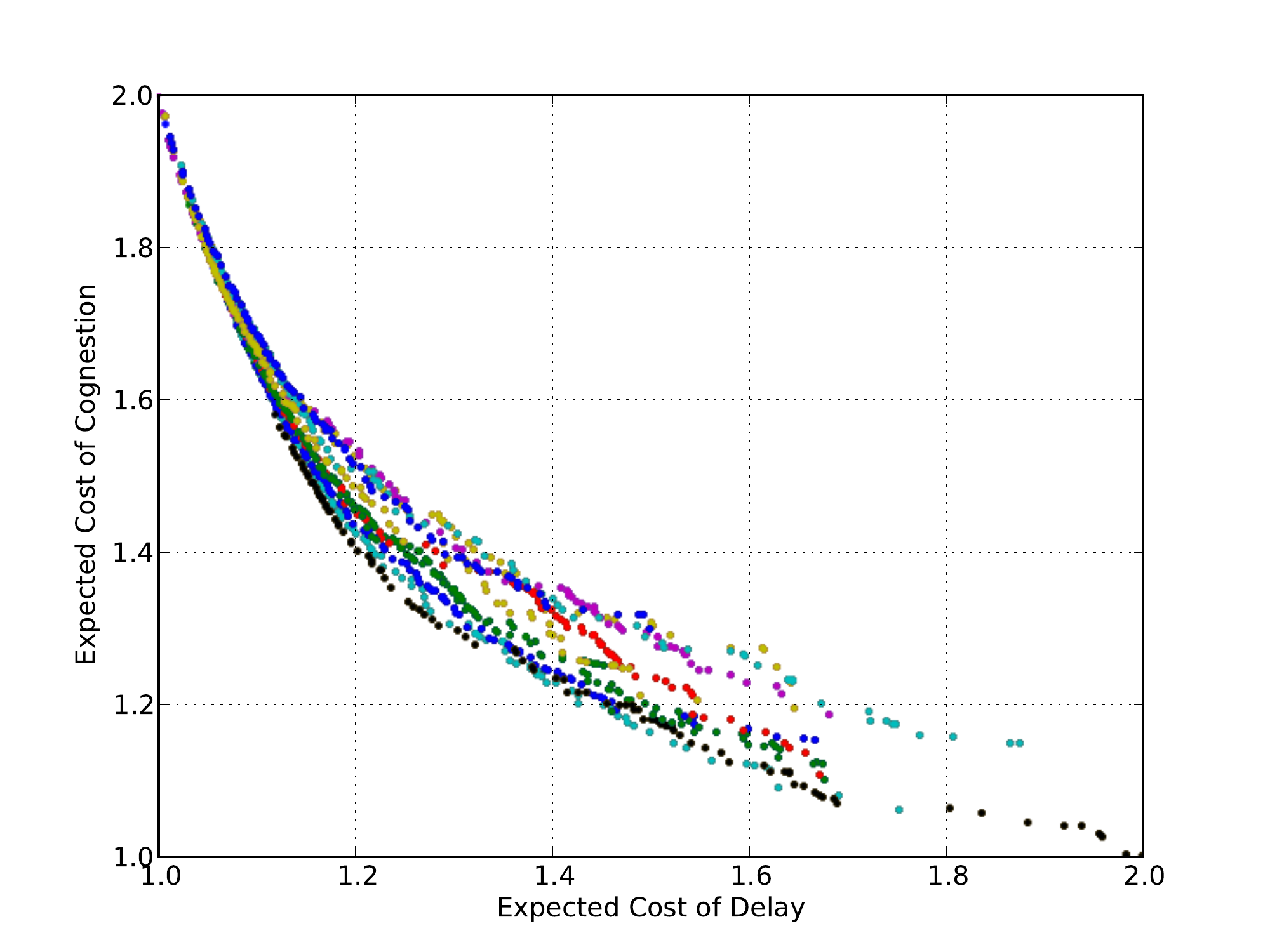}
  \caption{\label{fig:pareto}Pareto Front}
\end{figure}

\begin{figure}
  \centering
  \includegraphics[scale=0.35]{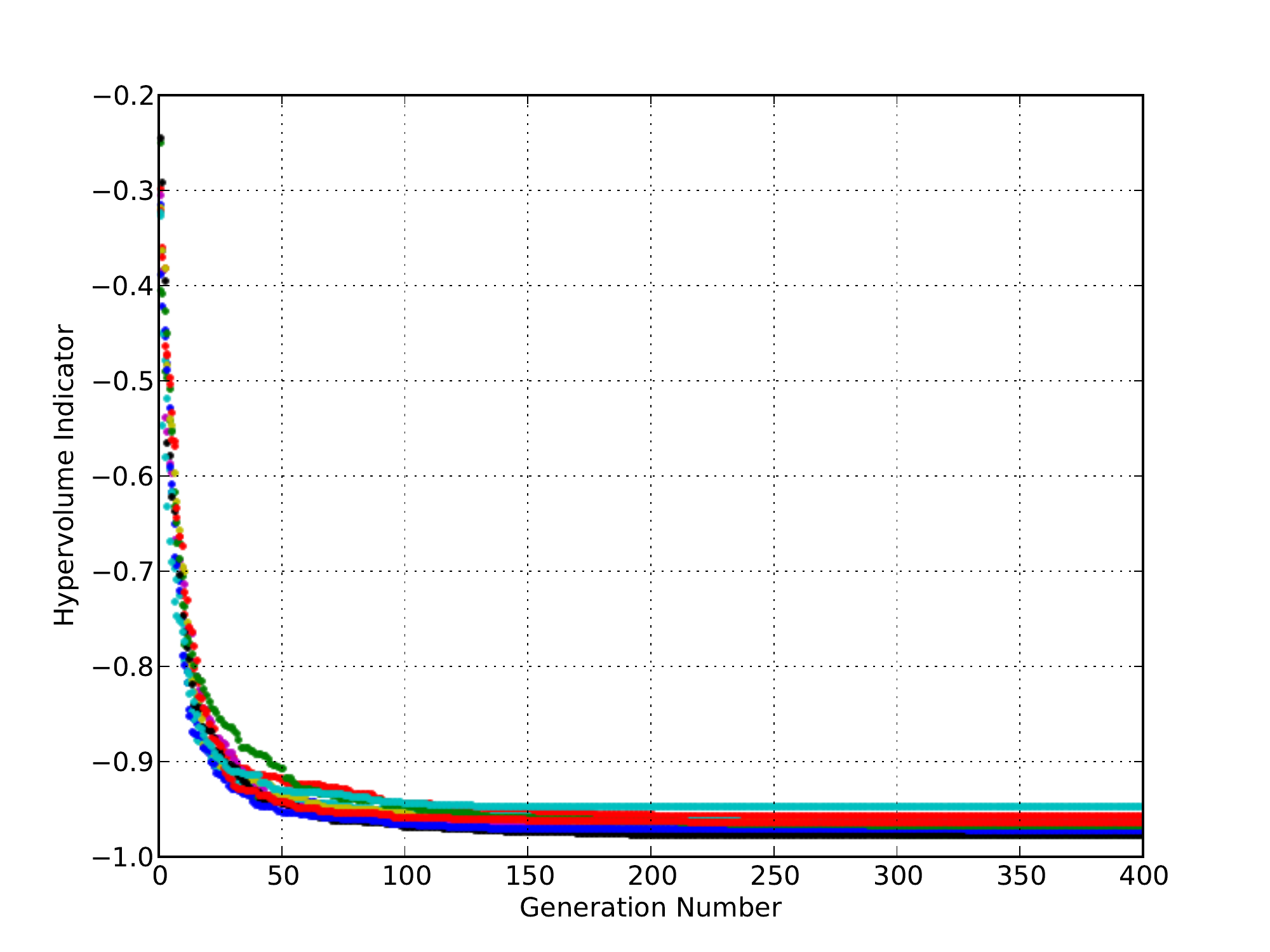}
  \caption{\label{fig:hypervolume}Hypervolume Indicator}
\end{figure}

Now, we can analyze the Pareto front obtained with the last population of 11 runs, depicted on figure \ref{fig:pareto}. 
At the first glance, the runs generate different fronts, which cover a large region, especially when minimizing the expected cost of congestion. 
So, minimizing the delays is relatively easy when we do not consider congestion. 
This is why the upper part of the Pareto fronts overlap. But, when we try to minimize the congestion, it seems that the algorithm falls in local optima. 
To verify this explanation, we used the hypervolume indicator, which gives the volume determined by the area enclosed by the Pareto front and the worst possible point. 
Figure \ref{fig:hypervolume} confirms the premature convergence since the hypervolume indicator stabilizes around generation 150 for each run.
In future work, we will tune the parameters of the algorithm according to the hypervolume indicator. 
We expect to find good parameters, maybe by simply increasing the exploration ratio, to avoid these local optima.

\section{Discussion}
One clear limit of this work is that the proposed theoretical framework has been validated on a single instance, assessing at the same time both the uncertainty model and the optimization process. 
Further experiments are mandatory to draw any firm conclusion regarding either the model or the optimization algorithm, and these experiments must involve several other instances.
Here, the second instance was created to assess the computation time of the probabilistic model.
We notice that the mean cost functions decrease during the optimization, but further investigation is necessary to understand how the algorithm impact decision variables that are not related to the disruption.
We think that the second instance is a good starting point to create different variant of disruptions and to assess that the approach is able to tackle the problem.

Even if the algorithm returns many solutions, a human operator will not be able to understand each of them in real-time. 
Consequently, the multiobjective algorithm shall return only a subset of interesting solutions. 
All the difficulty comes from the definition of what is an interesting solution. 
One way to circumvent this difficulty is to use the diversity operator used in several population-based multiobjective algorithms. 
A diversity operator is used to filter solutions, but at the same time, to preserve the shape of the pareto front. 
This prevents from having solutions in the same region. 
A necessary parameter is the size of the archive containing the solutions. 
Here, we used the crowding distance measure, inherent to NSGA-II with an archive size of 100 solutions.

The triangular distribution is an interesting choice because its variance increases as the mode approaches the bounds. 
In our case, the cost function is very similar to the variance formula, the target replaces the mean and we take only the positive part. 
Hence, we verified experimentally that the cost function increases as the target approaches the boundaries. 
This is relevant for the optimization algorithm since we penalize any target that would be close to the boundaries of the feasible set.

Regarding the optimization algorithm, there is very little hope to ever formally prove its convergence. 
Hence methods from experimental sciences must be used here. 
Statistics over many random instances of the size of a real operational context are the way to go, assessing how often and in which contexts the method can fails. 
This validation method will also provide insights on the actual computation burden that is required for large-scale instances. 

The choice of NSGA-II was motivated by the fact that we still do not fully understand all the properties underlying the probabilistic model. 
Hence, instead of trying to guess possibly wrong assumptions, we have chosen to use a robust and general optimization algorithm. 
Nevertheless, it is important in the future to compare NSGA-II with other, more recent multi-objective optimization algorithms (e.g., as already mentioned, MO-CMA-ES \cite{igel2007ecj}). 
Therefore, an extensive statistical study is needed, in order to find the most suited algorithm for this kind of problem.


Moreover, the formulation of the tactical planning problem shall also be extended to include more operational constraints.
Also, experiments and data mining shall be done in an operational context in order to model more accurately the underlying uncertainty.
The uncertainty could represent the expected errors of the trajectory prediction.
This would create a smooth transition between both phases.
Since we are in an online context, particle filters could also be used at the trajectory level in order to estimate the probability to be at the next points and therefore, estimate the probability of congestion.
With recent statistical studies on real trajectories, we know that the uncertainty is different from one phase to another. 
Indeed, there is important uncertainty on the real departure time and on the climbing phase. 
Afterward, the uncertainty is low in en-route phase and so, the values used here are certainly too high. 
In the following, we should configure the model to reflect the uncertainty for the different phases. 

Finally, we have mentioned the use of stochastic processes for the probability that the flight is in a sector and for the probability of congestion. We believe that there are interesting issues that can be addressed with the novel techniques from this domain.

\section{Conclusion}
This article has introduced a probabilistic model to handle the propagation of the uncertainty from the trajectories to the sectors.
The prerequisites of the model are the marginal of the initial arrival in the airspace, the policy of the flight management system when trying to stick to the target schedule, and the potential external disruptions. 
Thereafter, the probability for a flight to be in a sector can be computed. 
From there on, the closed-form equation to compute the probability of congestion, can be derived. 

Then, some general formulations for the expected cost of delays and the expected cost of congestion were given. We used a well-known trick to ensure equity that was naturally integrated in the model. Finally, because the congestion measure is clearly not the only criterion that should be used to decide for a schedule, the well-known multiobjective algorithm NSGA-II was proposed to solve the bi-objective problem of minimizing both the congestion and the cumulated delays of the flights, i.e., to approximate the non-dominated solutions of the Pareto front. 
These solutions are then proposed as alternatives for the decision maker, namely here the multi-sector planner.

Furthermore, in order to illustrate how the theoretical model can be useful in practice, we presented some results on two instances.
The results were analyzed to discover the consequences of some previous assumptions and to assess of the computation time of the inference mechanism.
One is the choice of triangular distributions, which is commonly used in project management.
On-going and further work will of course investigate other MOEAs to replace NSGA-II, and, more importantly, several different instances. One crucial issue is how well (or bad) this algorithm scales with the problem complexity (number of flights and number of sectors).
Nevertheless, we are confident that further studies will demonstrate the robustness of the proposed approach of using multiobjective evolutionary algorithms to solve the stochastic and dynamic optimization problem of air traffic flow and capacity management.

\section{Acknowledgments}
Ga{\'e}tan Marceau is funded by the scholarship CIFRE 710/2012 established between Thales Air Systems and the TAO project-team at INRIA Saclay Ile-de-France, and the scholarship 167544/2012-2013 from the {\it Fonds de Recherche du Qu{\'e}bec - Nature et Technologies}.


\bibliographystyle{IEEEtran}
\bibliography{library}

\end{document}